# WRICNet:A Weighted Rich-scale Inception Coder Network for Multi-Resolution Remote Sensing Image Change Detection

Yu Jiang, Lei Hu, Yongmei Zhang, and Xin Yang

*Abstract*—Majority models of remote sensing image changing detection can only get great effect in a specific resolution data set. With the purpose of improving change detection effectiveness of the model in the multi-resolution data set, a weighted rich-scale inception coder network (WRICNet) is proposed in this article, which can make a great fusion of shallow multi-scale features, and deep multi-scale features. The weighted rich-scale inception module of the proposed can obtain shallow multi-scale features, the weighted rich-scale coder module can obtain deep multi-scale features. The weighted scale block assigns appropriate weights to features of different scales, which can strengthen expressive ability of the edge of the changing area. The performance experiments on the multi-resolution data set demonstrate that, compared to the comparative methods, the proposed can further reduce the false alarm outside the change area, and the missed alarm in the change area, besides, the edge of the change area is more accurate. The ablation study of the proposed shows that the training strategy, and improvements of this article can improve the effectiveness of change detection.

*Index Terms*—Remote sensing(RS), change detection(CD), deep learning(DL), convolution neural network(CNN), rich-scale.

## I. INTRODUCTION

CHANGE detection(CD) can be defined as the identification of the differences in the state of an object or a pattern by observing at different times [1]. It can be applied in diverse applications, for instance, urbanization [2], monitoring of forest [3], water [4], shifting land use [5], and disaster [6]. Changes are high-lighted by means of some temporal indicators which are then processed for the extraction of meaningful changing patterns [7].

This work was supported by the National Natural Science Foundation of China under Grant No. 61662033.

Yu Jiang is with the the School of Computer Information Engineering, Jiangxi Normal University, Nanchang 330022, China.

Lei Hu is with the the School of Computer Information Engineering, Jiangxi Normal University, Nanchang 330022, China (e-mail: hulei@jxnu.edu.cn).

Yongmei Zhang is with the the School of Information, North China University of Technology, Beijing 100144, China.

Xin Yang is with the the School of Computer Information Engineering, Jiangxi Normal University, Nanchang 330022, China.

According to the analyzed object, CD can be divided into pixel-based [8]-[11], object-based [12], [13], and scene-based [14] CD. The traditional CD methods are supporting vector machine(SVM) [15], extreme learning machine(ELM) [16], markov random field(MRF) [17], conditional random field(CRF) [18], and random forest(RF) [19] *et al*. With the development of deep learning, neural network models such as convolutional neural network(CNN) [11], [20-22], recurrent neural network(RNN) [23], [24], generative adversarial network(GAN) [25], [26] have been successfully applied to CD.

Recently, since convolutional neural networks(CNNs) do well in extracting image features, plenty of classic CNNs, and improved models are used for CD, such as, CaffeNet [27], SegNet [28], U-net [18], [29] *et al*. Due to the difference in the size of the change area in CD, it is particularly important to represent the features of different scales. As expected, multi-scale features have been widely applied in feature extraction module. There are multiple ways to obtain multi-scale features, for instance, AlexNet [30], and VGGNet [31] obtains the multi-scale features by stacking convolutional operators, InceptionNets [32]-[35] use convolution layers with different kernel size, ResNet [36] use residual block, DenseNet [37] use shortcut connections, DLA [38] use hierarchical layer aggregation, and Rich-scale block [39] construct hierarchical residual-like connections within single residual block, which makes obtain multi-scale feature feasible.

Due to the lack of satisfactory features that can effectively distinguish the change area, the features are usually sensitive to factors such as noise, angle, shadow, and context, so the method is low robustness to pseudo-changes [40]. The attention mechanism(AM) is proved to be effective in solving this problem, for instance, the self-attention mechanism (Self-AM) [41] explores the performances of non-local operations of images, and videos in the space-time dimension, which is of substantial significance for the long-range dependencies of images. The Spatial Attention Mechanism (SAM) [42] can enhances the feature representations by encoding the context information of a long range into local features. Besides, the Channel Attention Mechanism(CAM) [43] can establish the relationships of features between channels. Certainly, the AM has been successfully applied to CD [40], [44].

With the development of Earth observation technology, more and more new satellite sensors are designed to acquire high-resolution(HR) remote sensing(RS) images, such as IKONOS, GaoFen(GF), SPOT, QuickBird *et al*. Compared with medium-resolution(MR), and low-resolution(LR) RS images, since HR RS images can provide abound spatial distribution information and surface details, CD in HR RS images is a popular research field. Nonetheless, different RS tasks require different resolutions of RS images, therefore, CD in multi-resolution RS images have great significance.

For the purpose of improving robustness of the model in the multi-resolution RS image, a WRICNet is designed in this article. First, we uses two branched to obtain shallow multi-scale features, and deep multi-scale features respectively, the former contains more details, but there are also some interference information, whereas the latter retains most of the important information, but discards some details. Second, the multi-scale feature extraction blocks are improved in this article, and a proposed block can assign appropriate weights to the features of different scales. In summary, the main contributions of this article are concluded as follows:

(1) Weighted scale block is proposed in this article. This structure can assign appropriate weights to the features of different scales, so as to more accurately represent the edge of the changing area, which is conducive to reducing the missing, and false alarms at the edge of the changing area.
(2) Rich-scale block is improved in this article. On the basis of not changing the scale obtained by each group, the independence of the feature of each group, and the expression ability of multi-scale features are strengthened.
(3) Inception module is improved in this article. Without changing the scale obtained by each branch, the model parameters are reduced.
(4) WRICNet is designed in this article, which can make a great fusion of shallow multi-scale features, and deep multi-scale features, so that the model has great robustness in multi-resolution RS images.

The rest of the article is organized as follows. The proposed methodology is presented in Section II. The data sets, experiment settings, experiment performance, and comparison against the comparative methods are discussed in Section III. At the end, we make a conclusion in Section IV.

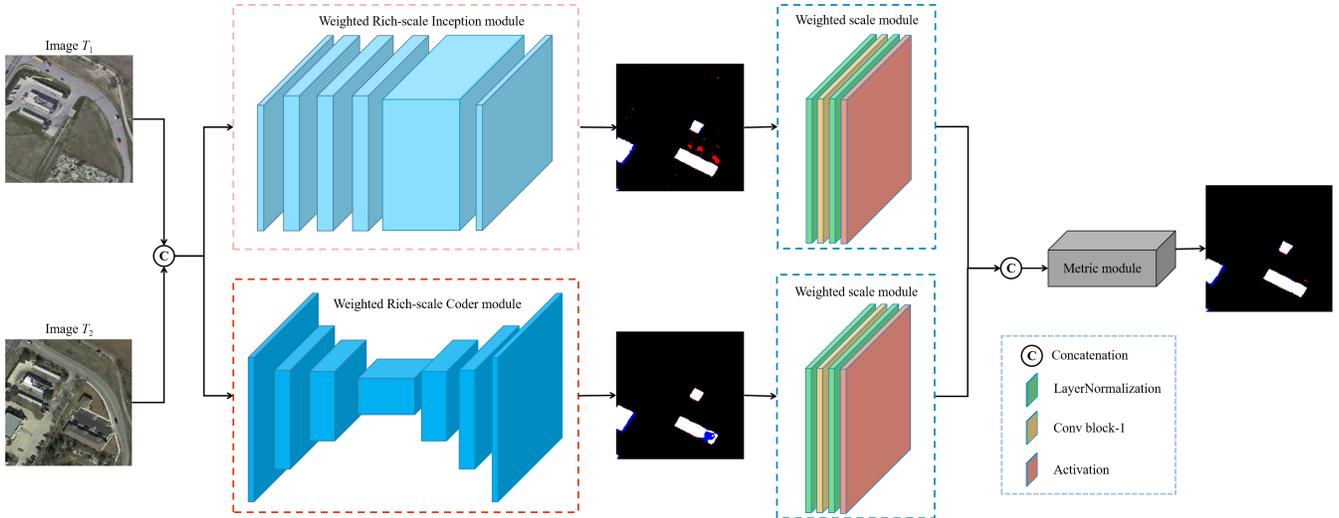

Fig. 1. Overview of the proposed Network. Black indicates the non-change detected correctly. Black indicates the non-change detected correctly. White indicates the change detected correctly. Red indicates the change detected by error. Blue indicates unpredicted change.

## II. METHODOLOGY

### A. Weighted Rich-scale Inception Coder Network

As illustrated in Fig. 1, the proposed(WRICNet) is an end-to-end network which takes the Image $T_1$, and Image $T_2$ as input, and makes a Concatenation between Image $T_1$, and Image $T_2$ in the channel dimension before feature extraction. It consists of 4 components: a weighted rich-scale inception(WRI) module, a weighted rich-coder coder(WRC) module, a Weighted scale block, and a Metric module.

The WRI module, and WRC module have the ability to obtain multi-scale features. The former does not change the size of the feature in the process of obtaining the multi-scale feature, besides, the input feature, and the output feature have the same dimension. Therefore, in this article, the obtained feature is called a shallow multi-scale feature(SMF), which has the characteristics of fewer missed alarms in the changing area, but more false alarms outside the changing area. The latter first reduces the feature size, and then enlarges the feature size in the process of obtaining multi-scale features, furthermore, the input feature, and the output feature have the same dimension. Thus, in this article, the obtained feature is called a deep multi-scale feature(DMF), which has the characteristics of more missed alarms in the changing area, but fewer false alarms outside the changing area. The Weighted scale block assigns appropriate weights to SMF, and DMF, and then fuses them via Concatenation. Finally, the Metric module with Densely connected structure [37] is used to obtain the CD output.

## B. Weighted Scale Block

As illustrated in Fig. 2, with the purpose of assigning appropriate weights to the multi-scale feature, the Weighted scale block is designed in this article. It consists of LayerNormalization, Conv block-1, LayerNormalization, Activation, and Multiply operating, where the Kernel_size, Kernel_initializer, Stride, Padding, Activation of Conv block-1 are set to 1, "ones", 1, "same", "sigmoid".

## C. Rich-scale Block

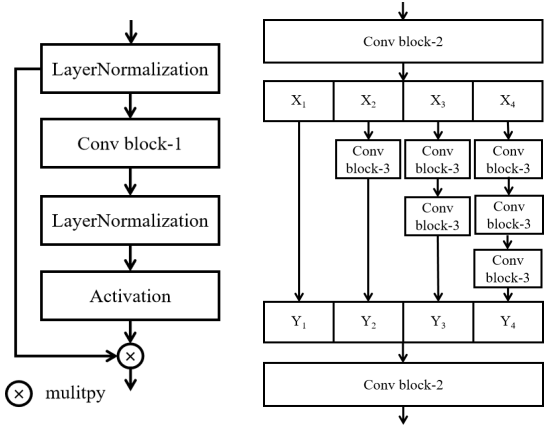

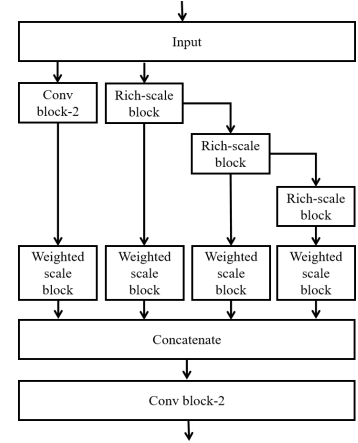

Fig. 4. Weighted Rich-scale Inception module.

Fig. 2. Weighted scale block    Fig. 3. Rich-scale block

To strengthen the ability of model to obtain multi-scale features, the literature [42] designs Rich-scale block, which can be plugged into the state-of-the-art backbone CNN models. The Rich-scale block divides the features into multiple groups in the channel dimension. The latter group fuses the features obtained by the previous group, and then makes feature extraction, so that each group can obtain different receptive field via different numbers of convolutions, so as to obtain multi-scale features.

Since the input feature of each group is different, without changing the scale obtained by each group, as illustrated in Fig. 3, this article uses different numbers of Conv block-3 to obtain multi-scale feature in each group rather than fuse the features obtained by the previous group, to a certain extent, which can strengthen the independence of each group feature, and the expression ability of multi-scale features. The Kernel_size, Stride, Padding, and Activation of Conv block-3 are set to 3, 1, "same", and "relu".

## D. Weighted Rich-scale Inception Module

Based on Hebbian theory introducing sparse features, and extracting multi-scale features by widening the network structure, Szegedy *et al.* [32]-[35] designs the Inception module. In this structure, each branch uses a different number of convolutions to obtain features of different scales [45].

Since the input of each branch is the same, the feature scales obtained after the same number of convolutions on each branch are the same. According to the prior knowledge, this article regards the extracted features of each branch, except the first branch, as input of the next branch. Therefore, without changing the scale obtained by each branch, and reduces the model parameters by reducing the number of convolutions in the branch.

As illustrated in Fig. 4, with the purpose of further improving the ability of extracting multi-scale features, this article designs WRI module which replaces the convolution block in the Inception module with the Rich-scale block. The Weighed scale block is used to assign different weights to the features extracted by each branch, and then fuses features of different scales via Concatenate. In the first branch, the Kernel_size, Stride, Padding, and Activation of Conv block-2 are set to 1, 1, "same", and "relu".

It's worth noting that WRI module does not change the size of the feature in the process of extracting the multi-scale feature, besides, the input feature, and the output feature have the same dimension.

## E. Weighted Rich-scale Coder Module

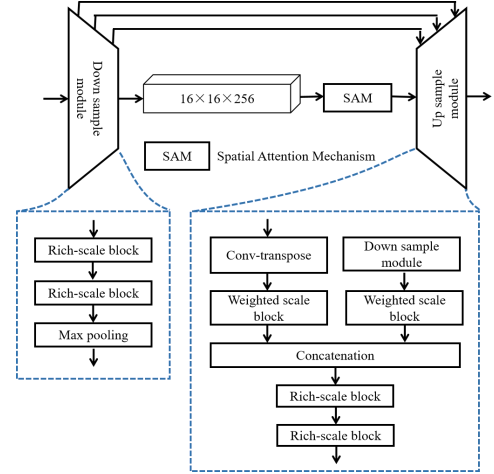

Fig. 5. Weighted Rich-scale Coder module

As illustrated in Fig. 5, this article designs WRC module, which based on U-net, and applies the Weighted scale block, the Rich-scale block to Down Sample(DS) module, and Up Sample(US) module. The DS module consists of 4 DS operations. Each DS operation consists of 2 Rich-scale block, and 1 Max pooling operation. After each DS operation, the channels of the features are 32, 64, 128, and 256. The dimension(W×H×C) of feature changed from 256×256×6 to 16 × 16 × 256, then via SAM [40] to obtain the long-range spatial-temporal attention. US module consists of 4 US operations. In each US operation, first, Conv-transpose the feature, before via Concatenate makes skip-connection with the feature of the same dimension in the DS module, via the Weighted scale block assigns appropriate weights to the both, then, extracts the multi-scale feature by 2 Rich-scale block. After each US operation, the channels of the features are 128, 64, 32, and 6.

It is worth noting that WRC module will change the size of the feature in the process of extracting the multi-scale feature, whereas the input feature, and the output feature have the same dimension.

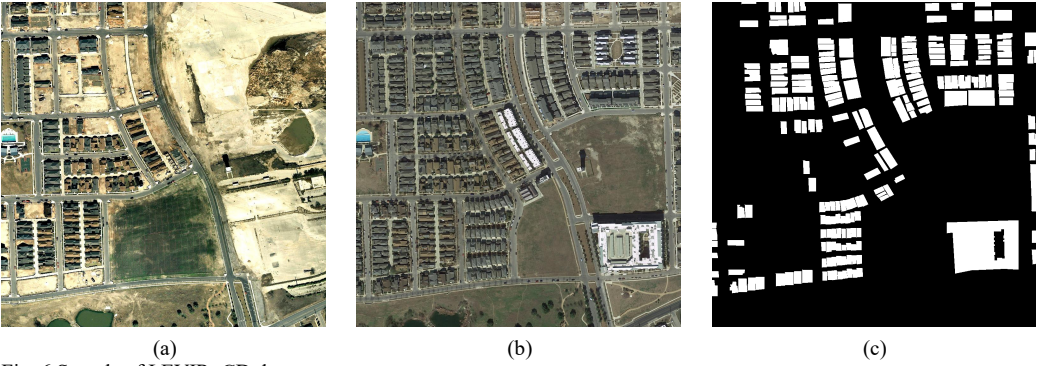
(a) (b) (c)
Fig. 6.Sample of LEVIR_CD data set.

## III. EXPERIMENTS

### A. Date set

The HR data set used in this article is LEVIR_CD [44], which is collected from Google Earth, and contains 637 pairs of HR images with a size of 1024 × 1024 pixels, and 0.5 meters per pixel of American cities in 2002, and 2018. As illustrated in Fig. 6, each pair of images consists of 3 images(Image T1, Image T2, and Ground Truth). This HR data set used for building CD.

In order to discuss the CD effect of the model in different resolution data sets, this article uses the BicubicInterpolation, and the Nearest Neighbor Interpolation to resize the Image $T_1$, Image $T_2$, and Ground Truth of the HR data set, respectively. The HR date set(1024 × 1024) reduces the resolution by 2 times to get the Medium-Resolution(MR) data set(512×512), and reduces the resolution by 4 times to get the Low-Resolution(LR) data set(256×256).

Since the size of the network input is 256×256, this article crops the HR set, and MR data set by a sliding window with a size of 256×256, and a stride of 256. In order to improve the robustness of the model, this article randomly selects a method for data enhancement. Among them, the data enhancement methods include up-down flip, left-right flip, and 90° counterclockwise rotation. In order to speed up the convergence of the model during the training process, min-max normalization is used for the data set, and the range is [0, 1].

### B. Implementation Details:

*1) Experimental Settings:* As illustrated in formula(1), the loss function used in this article is the joint loss, which consists of $L_{ri}$, $L_{wri}$, $L_{ed}$, $L_{wed}$ and $L_{fu}$. As shown in formulas(2)-(6), the method of calculating each loss is cross-entropy loss.

$$Loss = \lambda_1 L_{ri} + \lambda_2 L_{wri} + \lambda_3 L_{ed} + \lambda_4 L_{wed} + \lambda_5 L_{fu} \quad (1)$$

$$L_{ri} = -\sum_{i=1}^{C} I_i^{gt} \cdot \log(I_i^{ri}) \quad (2)$$

$$L_{wri} = -\sum_{i=1}^{C} I_i^{gt} \cdot \log(I_i^{wri}) \quad (3)$$

$$L_{ed} = -\sum_{i=1}^{C} I_i^{gt} \cdot \log(I_i^{ed}) \quad (4)$$

$$L_{wed} = -\sum_{i=1}^{C} I_i^{gt} \cdot \log(I_i^{wed}) \quad (5)$$

$$L_{fu} = -\sum_{i=1}^{C} I_i^{gt} \cdot \log(I_i^{fu}) \quad (6)$$

Where $C$ is the category, which is set to 2 in this article, $I^{gt}$ is the Ground Truth, $I^{ri}$ is output of WRI module, $I^{wri}$ is the weighted map obtained from $I^{ri}$ and Weighted scale block. $I^{ed}$ is output of WRC module, $I^{wed}$ is the weighted map obtained from $I^{ed}$ and Weighted scale block. $I^{fu}$ is the CD output obtained by fusing SMF, and DMF. $L_{ri}$ is the cross entropy loss between $I^{ri}$, and $I^{gt}$. $L_{wri}$ is the cross entropy loss between $I^{wri}$, and $I^{gt}$. $L_{ed}$ is the cross entropy loss between $I^{ed}$, and $I^{gt}$. $L_{wed}$ is the cross entropy loss between $I^{wed}$, and $I^{gt}$. $L_{fu}$ is the cross entropy loss between $I^{fu}$, and $I^{gt}$.

Moreover, the proposed is implement based on the Tensorflow framework, and a single NVIDIA GTX 1080TI GPU with 11G memory is used for training, and testing. The epoch, and batch size are set to 200, and 1. In this article, Adam is used as an optimizer, and the initial learning rate, $\beta_1$, and $\beta_2$ are set to 1e-4, 0.9, and 0.999. The $\lambda_1$, $\lambda_2$, $\lambda_3$, $\lambda_4$, and $\lambda_5$ of formula(1) are set to 1, 1.3, 0.5, 0.65, and 2.3.

$$\begin{cases} c\_sample = \sum (img_{i,j} = 1., 0 \leq i, j \leq 255) \\ uc\_sample = \sum (img_{i,j} = 0., 0 \leq i, j \leq 255) \\ c\_weight = 1. \\ uc\_weight = uc\_sample / c\_sample. \end{cases} \quad (7)$$

Since the imbalance in the number of change, and non-change samples in the data set, therefore, as illustrated in formula(7), the proposed can automatically count the number of change, and non-change samples in the data set during the training process, in addition, assigns appropriate weights to the change, and non-change samples according to the statistical results.

Where $c\_sample$, and $uc\_sample$ are the number of change samples, and non-change samples, respectively. $img_{i,j}$ is the value of the normalized $img$ at coordinates(i, j), $c\_weight$ is the weight of the changed samples, $uc\_weight$ is the weight of the non-change samples.

*2) Comparative Method:* To verify the effectiveness of the proposed(WRICNet), some CD methods are compared, and analyzed, which are as follows:

(1) CDNet [8]: A deep Deconvolutional Network for pixel-based street-view CD, which based on ideal of stacking contraction, and expansion blocks.
(2) U-net [9], [29]: A U-net network with early fusion for CD, which only contains 4 max pooling, and 4 up sample operations, thus, the layers are shallower than traditional U-net equivalents.
(3) FC-Siam-conc [9]: A fully convolutional network with Siamese extension of U-net model for CD, which is combination of a fully convolutional encoder-decoder paradigm and a siamese architecture, and concatenates the skip connections during the decoding steps.
(4) FC-Siam-diff [9]: A siamese extension of fully convolutional network with difference for CD, which concatenates the absolute value of their difference from the encoding steams, instead of concatenating both connections.
(5) STANet [44]: A spatial-temporal attention neural network for remote sensing image binary building CD, and use SAM capture the long-range spatial-temporal attention, and obtain the multi-scale attention features, which is more robust to color, and scale variations in bitemporal images

*3) Evaluation Metrics:* Miss Alarm(MA), False Alarm(FA), F1-Score(F1), Intersection over Union(IoU) are used to quantitatively evaluate the performance of the comparative methods and the proposed. These metrics are calculated as follows:

$$MA = 1 - \frac{TP}{TP + FN} \quad (8)$$

$$FA = \frac{FP}{FP + TP} \quad (9)$$

$$\begin{cases} P = \dfrac{TP}{TP + FP} \\ R = \dfrac{TP}{TP + FN} \\ F1 = \dfrac{2PR}{P + R} \end{cases} \quad (10)$$

$$IoU = \frac{TP}{TP + FP + FN} \quad (11)$$

Where False Position(FP) is the quantity of non-change pixels unpredicted, True Positive(TP) is the quantity of non-change pixels detected correctly, False Negative(FN) is the quantity of change pixels unpredicted, and True Negative(TN) is the quantity of change pixels detected correctly. Among them, the F1, and IoU can better reflect the generalization ability of a model.

MA can intuitively reflect the probability of the true change in all pixels but detected non-changed, FA can intuitively reflect the probability of the true non-change in all pixels but detected changed, the lower the probability, the better the CD effect. F1 is the harmonic mean of $P$ and $R$. The IoU is the area of overlap between the predicted changed pixels, and the changed pixels divided by the union area between them. The F1, and IoU can comprehensively reflect the CD effectiveness of a model, and the higher the probability, the better the CD effect.

To evaluate the CD effect, this article calculates the Local Optimal Index(LOI), and the Global Index(GI). According to the descending order of F1 in the CD output, MA, FA, F1, and IoU are calculated as the LOI based on the top 5%, and 10% of the confusion matrix. Calculating MA, FA, F1, IoU as GI based on the total confusion matrix.

*C. Performance Experiment*

*1) Performance on high, medium, and low-resolution data set*

The CD output of comparative methods, and the proposed are illustrated in Fig. 7. Through careful observation, we can find that there are a lot of FA, MA in the CD output of CDNet(d), the former majority located at at the edge of the change area, the later majority located at at the larger change area. Since CDNet can obtain large-scale features, compared to the MR, and LR data set, the CD effect of output on HR data set is better. Compared with CD output of CDNet, U-net(e), FC-Siam-conc(f), and FC-Siam-diff(g) can obtain multi-scale features, which reduce MA, and FA of CD output to a certain extent. But MA abound in the larger change area, and a certain amount of FA still exists at the edge of the change area. The worth mentioning is that compared to the CD output of U-net, and FC-Siam-diff, the output of FC-Siam-conc has less MA, and FA, besides, the MA is the least in the larger change area. Since STANe(h)t can obtain multi-scale features, and the spatial mapping relationship between features. Therefore, STANet can more correctly detect the change area, which reduce the MA in the change, but there are lots of FA at the edge of the change area. Compared with the comparative methods, in the proposed(i) CD output, The MA is the least in the large change area, the FA outside the change area, and the edge of the change area is the least, all in all, CD output of the proposed is the closest to Ground Truth.

To quantitatively analyze the effect, top 5%, 10% LOI, GI of comparative methods, the proposed on HR, MR, and LR data set are calculated, which summarized in Table I, Table II, and Table III, respectively.

As illustrated in Table I, the proposed obtains the best FA, F1, and IoU of the LOI, and GI on the HR data set. In addition, it is worth noting that STANet obtained the best MA of LOI, and GI on the HR dataset, F1, and IoU of LOI are relatively close to U-net, and obtained better GI(F1, IoU) than FC-Siam-conc.

As illustrated in Table II, the proposed obtains the best MA, FA, F1, and IoU of the LOI, and GI on the MR data set. Besides, it is worth mentioning that FC-Siam-diff obtains the best MA, FA, F1, and IoU of the LOI, and GI in comparative methods.

As illustrated in Table III, the proposed obtains the best FA, F1, and IoU of the LOI, and the best MA, FA, F1, and IoU of GI on the LR data set. In addition, it is worth mentioning that STANet obtains the best MA of the LOI.

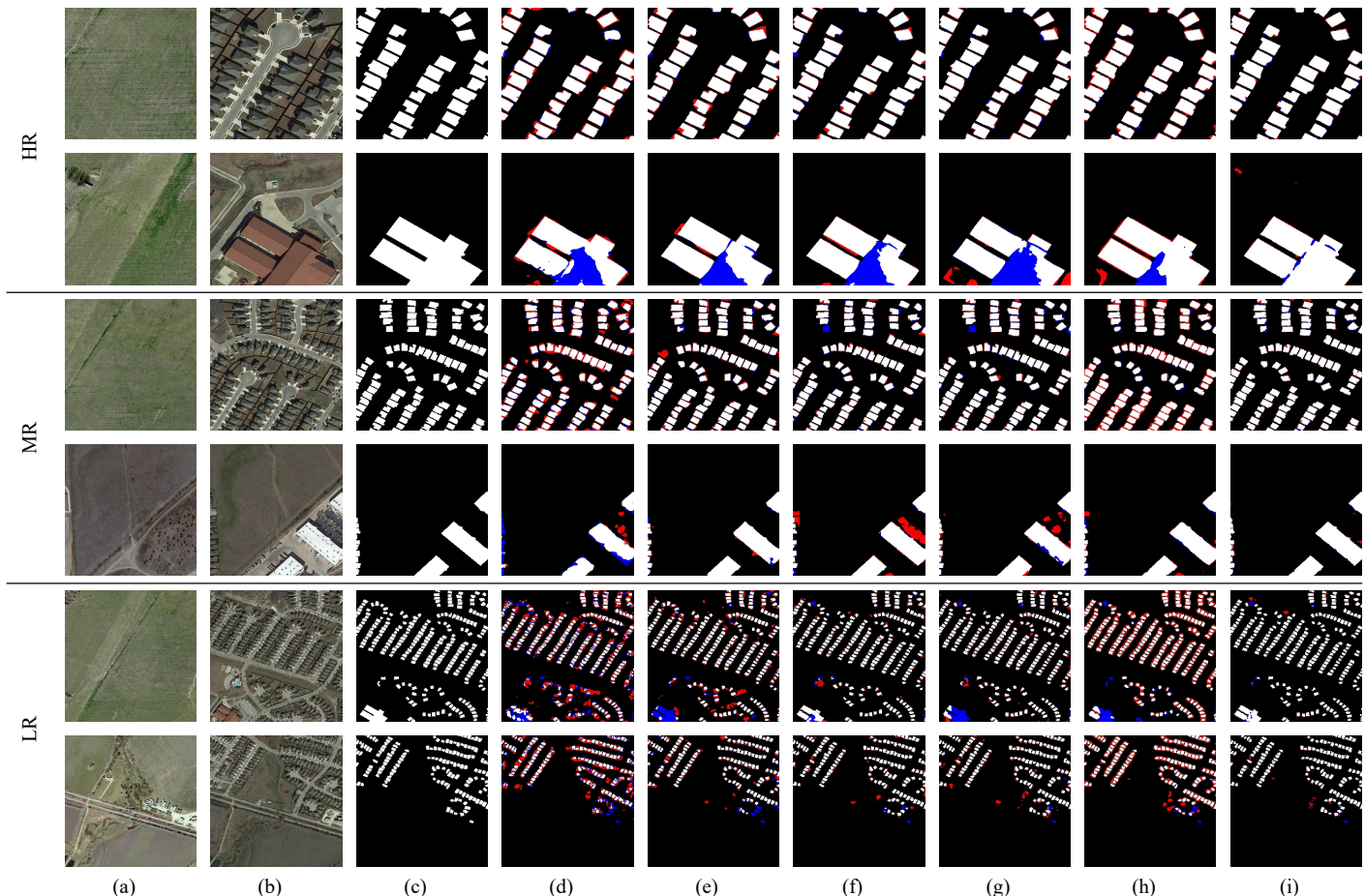

Fig. 7. Performance on medium-resolution data set. (a) Image $T_1$. (b) Image $T_2$. (c) Ground Truth. (d) CDNet. (e) U-net. (f) FC-Siam-conc. (g) FC-Siam-diff. (h) STANet. (i) Proposed(WRICNet).

TABLE I
PERFORMANCE ON HIGH RESOLUTION DATA SET

| Method | Local Optimal Index | | | | | | | | Global Index | | | |
|---|---|---|---|---|---|---|---|---|---|---|---|---|
| | 5% | | | | 10% | | | | | | | |
| | MA | FA | F1 | IoU | MA | FA | F1 | IoU | MA | FA | F1 | IoU |
| CDNet | 8.5 | 8.8 | 91.3 | 84.1 | 9.7 | 9.9 | 90.2 | 82.2 | 31.9 | 14.3 | 75.9 | 62.1 |
| U-net | 4.5 | 6.1 | 94.7 | 89.9 | 5.7 | 6.8 | 93.8 | 88.2 | 15.7 | 12.7 | 85.8 | 75.1 |
| FC-Siam-conc | 4.4 | 5.3 | 95.1 | 90.7 | 5.7 | 6.1 | 94.1 | 88.9 | 13.4 | 10.7 | 87.9 | 78.4 |
| FC-Siam-diff | 4.8 | 5.0 | 95.1 | 90.6 | 6.0 | 5.7 | 94.1 | 88.9 | 14.5 | 10.1 | 87.6 | 78.0 |
| STANet | **3.5** | 7.4 | 94.5 | 89.6 | **3.8** | 8.9 | 93.6 | 87.9 | **8.6** | 15.0 | 88.1 | 78.7 |
| Proposed | 4.9 | **4.8** | **95.2** | **90.8** | 6.0 | **5.3** | **94.3** | **89.3** | 13.9 | **9.6** | **88.2** | **78.9** |

TABLE II
PERFORMANCE ON MEDIUM-RESOLUTION DATA SET

| Method | Local Optimal Index | | | | | | | | Global Index | | | |
|---|---|---|---|---|---|---|---|---|---|---|---|---|
| | 5% | | | | 10% | | | | | | | |
| | MA | FA | F1 | IoU | MA | FA | F1 | IoU | MA | FA | F1 | IoU |
| CDNet | 13.5 | 14.9 | 85.8 | 75.1 | 14.7 | 17.1 | 84.1 | 72.6 | 36.1 | 22.7 | 69.9 | 53.8 |
| U-net | 7.3 | 8.1 | 92.3 | 85.7 | 8.5 | 8.6 | 91.4 | 84.2 | 22.2 | 14.9 | 81.3 | 68.4 |
| FC-Siam-conc | 6.9 | 6.7 | 93.2 | 87.2 | 8.2 | 7.4 | 92.2 | 85.5 | 18.6 | 13.4 | 83.9 | 72.2 |
| FC-Siam-diff | 6.3 | 7.1 | 93.3 | 87.4 | 8.1 | 7.5 | 92.2 | 85.5 | 18.7 | 12.9 | 84.1 | 72.6 |
| STANet | 7.8 | 12.8 | 89.6 | 81.2 | 8.1 | 14.7 | 88.5 | 79.3 | 27.4 | 17.3 | 77.3 | 63.0 |
| Proposed | **5.8** | **5.8** | **94.2** | **89.1** | **7.0** | **5.9** | **93.5** | **87.9** | **18.3** | **10.2** | **85.5** | **74.7** |

TABLE III
PERFORMANCE ON LOW-RESOLUTION DATA SET

| Method | Local Optimal Index | | | | | | | | Global Index | | | |
|---|---|---|---|---|---|---|---|---|---|---|---|---|
| | 5% | | | | 10% | | | | | | | |
| | MA | FA | F1 | IoU | MA | FA | F1 | IoU | MA | FA | F1 | IoU |
| CDNet | 26.3 | 26.9 | 73.4 | 58.0 | 28.5 | 30.6 | 70.4 | 54.3 | 50.7 | 40.7 | 53.8 | 36.8 |
| U-net | 16.8 | 14.7 | 84.3 | 72.8 | 17.7 | 17.4 | 82.5 | 70.2 | 38.3 | 24.1 | 68.1 | 51.6 |
| FC-Siam-conc | 9.9 | 8.9 | 90.6 | 82.8 | 10.5 | 11.2 | 89.2 | 80.4 | 23.1 | 18.4 | 79.2 | 65.6 |
| FC-Siam-diff | 11.6 | 10.3 | 89.0 | 80.3 | 11.7 | 12.4 | 88.0 | 78.6 | 26.7 | 19.3 | 76.8 | 62.3 |
| STANet | **6.5** | 22.3 | 84.9 | 73.8 | **8.3** | 23.9 | 83.2 | 71.2 | **21.4** | 32.6 | 72.6 | 56.9 |
| Proposed | 7.4 | **7.1** | **92.8** | **86.5** | 9.0 | **7.0** | **92.0** | **85.2** | 22.7 | **11.2** | **82.6** | **70.4** |

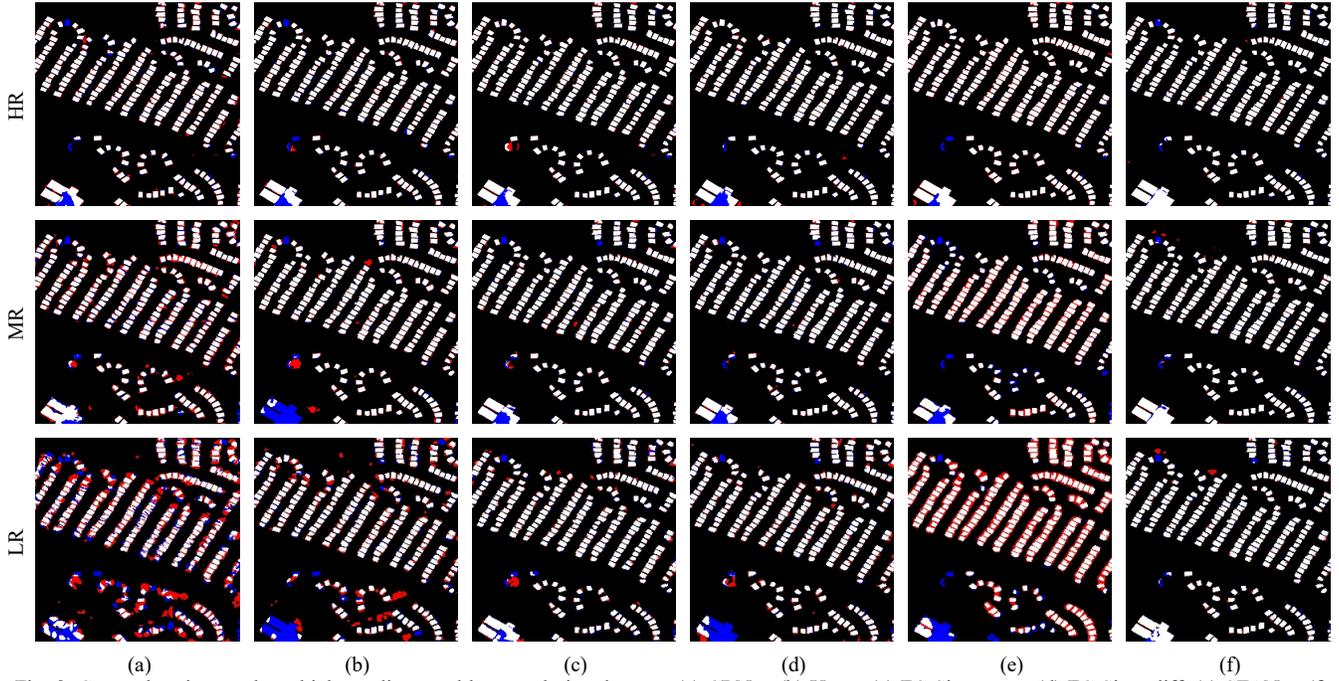

Fig. 8. Comprehensive result on high, medium, and low-resolution data set. (a) CDNet. (b) U-net. (c) FC-Siam-conc. (d) FC-Siam-diff. (e) STANet. (f) Proposed(WRICNet).

*2) Comprehensive Performance on high, medium, and low-resolution data set*

With the purpose of further comprehensively analyzing the overall CD effect of comparison methods, and the proposed on HR, MR, and LR data set, as illustrated in Fig. 8, this article restores the cropped change detection output on HR, and MR data set into a complete image, and compares it with the output on the LR data set. It should be noted that the Image $T_1$, Image $T_2$, and Ground Truth are the same as images in columns 1-3 of row 5 in Fig. 7.

Through careful observation, as the resolution of the data set decreases, Since CDNet(a) can only obtain large-scale features, the MA in the change area, the FA outside the change area, and the FA at the edge gradually increase, and the increase is the largest. U-net(b), FC-Siam-conc(c), and FC-Siam-diff(d) can obtain multi-scale features, therefore, the increase of MA, and FA is smaller than that of CDNet, and the increase of FC-Siam-conc is the smallest. Since STANet(e) can use SAM capture the long-range spatial-temporal attention, and obtain the multi-scale attention features, so there are fewer MA in the output, but the FA at the edge of the change area gradually increase. Compared with the comparative methods, the CD output of the proposed has the smallest increase of MA, and FA. The MA of the comparative methods in the larger change area gradually increases whilst the size of the change area differs greatly, since the proposed uses the Weighted scale block to assign weights to features of different scales, so the increase in MA is the least.

All in all, we can draw a conclusion that the output of the proposed on HR, MR, and LR data set is the closest to Ground Truth, and as the resolution of the data set decreases, the gap of the CD effect between comparison methods, and the proposed gradually increases. The above proves that the proposed has strong robustness to multi-resolution data sets.

To analyze the parameters, LOI, and GI of comparative methods and the proposed, model parameters of comparative methods, and the proposed model(a), LOI, and GI of comparative methods, and the proposed(b), (c) are illustrated in Fig. 9. It is worth noting that LOI, and GI are obtained by calculating the average value on HR, MR, and LR data set.

As illustrated in Fig. 9(a), through careful observation, we can find parameters of the proposed are 178%, 128%, 84%, 102%, and 17% of comparative methods, respectively. Among them, parameters of CDNet are the least, on contrast, since STANet uses ResNet as a pre-training model, it has the most model parameters. The model parameters of the proposed are closest to FC-Siam-diff.

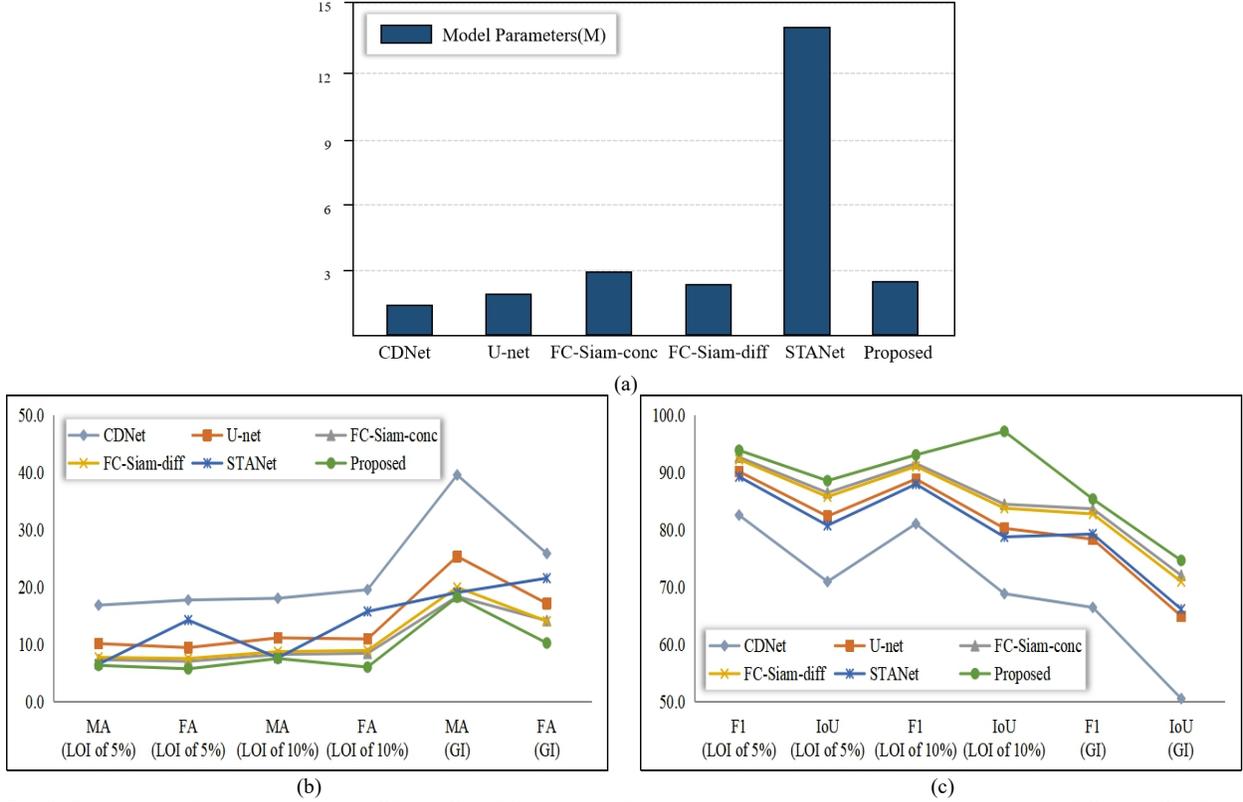

Fig. 9. Comparison of model parameters, LOI, and GI. (a) Parameters of comparative methods, and the proposed model. (b) LOI, and GI of MA, FA of comparative methods, and the proposed. (c) LOI, and GI of F1, IoU of comparative methods, and the proposed.

As illustrated in Fig. 9(b), the proposed obtains the optimal MA, and FA, which are 6.4%, 7.6%, 18.3%, and 5.8%, 6.1%, 10.3% of 5%, 10% LOI, and GI, respectively. Compared with comparative methods, for the MA of 5% LOI, the proposed reduces 10.5%, 3.8%, 1.0%, 1.4%, and 0.2%, respectively. For the FA of 5% LOI, the proposed reduces 12.0%, 3.7%, 1.3%, 1.8%, and 8.5%, respectively. For the MA of 10% LOI, the proposed reduces 10.5%, 3.6%, 0.7%, 1.2%, and 1.1%, respectively. For the FA of 10% LOI, the proposed reduces 13.5%, 4.9%, 2.4%, 2.9%, and 9.7%, respectively. For the MA of GI, the proposed reduces 21.3%, 7.1%, 0.1%, 1.7%, and 0.8%, respectively. For the FA of GI, the proposed reduces 15.6%, 6.9%, 3.9%, 3.8%, and 11.3%, respectively.

As illustrated in Fig. 9(c), the proposed obtains the optimal F1, and IOU, which are 93.9%, 93.1%, 85.4%, and 88.6%, 87.2%, 74.4% of 5%, 10% LOI, and GI, respectively. Compared with comparative methods, for the F1 of 5% LOI, the proposed improves 11.3%, 3.7%, 1.2%, 1.6%, and 4.6%, respectively. For the IoU of 5% LOI, the proposed improves 17.6%, 6.2%, 2.1%, 2.8%, and 7.8%, respectively. For the F1 of 10% LOI, the proposed improves 12.0%, 4.2%, 1.5%, 2.0%, and 1.8%, respectively. For the IoU of 10% LOI, the proposed improves 18.3%, 6.9%, 2.7%, 3.4%, and 8.4%, respectively. For the F1 of GI, the proposed improves 18.9%, 7.0%, 1.7%, 2.6%, and 6.1%, respectively. For the IoU of GI, the proposed reduces 23.8%, 9.4%, 2.3%, 3.4%, and 8.2%, respectively.

In terms of changing trend of comparative methods, and the proposed of LOI, and GI, we can draw a conclusion that the changing trend of the proposed with the minimum variation. The above proves that the proposed has fewer model parameters, and has strong robustness to multi-resolution data sets.

## D. Comparison of the Output of WRICNet

To verify the effectiveness of WRI module to obtain the shallow multi-scale feature, the effectiveness of WRC module to obtain the deep multi-scale feature, and the ability of the weighted scale block to express the edge of changing area, as illustrated in Fig. 10, this article compares the heat map of weighted output of WRI module, WRC module, and output of WRI module, WRC module, WRICNet on HR, MR, and LR data set.

Through careful observation, we can find that some of the following rules:

(1) In the output of the WRI module(f), there are fewer MA in the changing area, but more FA outside the changing area.
(2) In the output of the WRC module(g), there are more MA in the changing area, but fewer FA outside the changing area.
(3) In the heat map of weighted output of WRI module(d), and weighted output of WRC module(e), the area with the greatest weight is mainly located at the edge of the changing area.
(4) Compared with the output of WRI module(f), and WRC module(g), the output of WRICNet(h) has fewer MA in the change area, fewer FA outside the change area, and the edges of the changing area are more precise. All in all, output of WRICNet(h) is closest to Ground Truth.

According to the above, it can be proved that WRI module, and WRC module can obtain SMF, and DMF respectively. Weighted scale block can make a great expression of the edge of the changing area, and WRICNet can complement SMF, and DMF well.

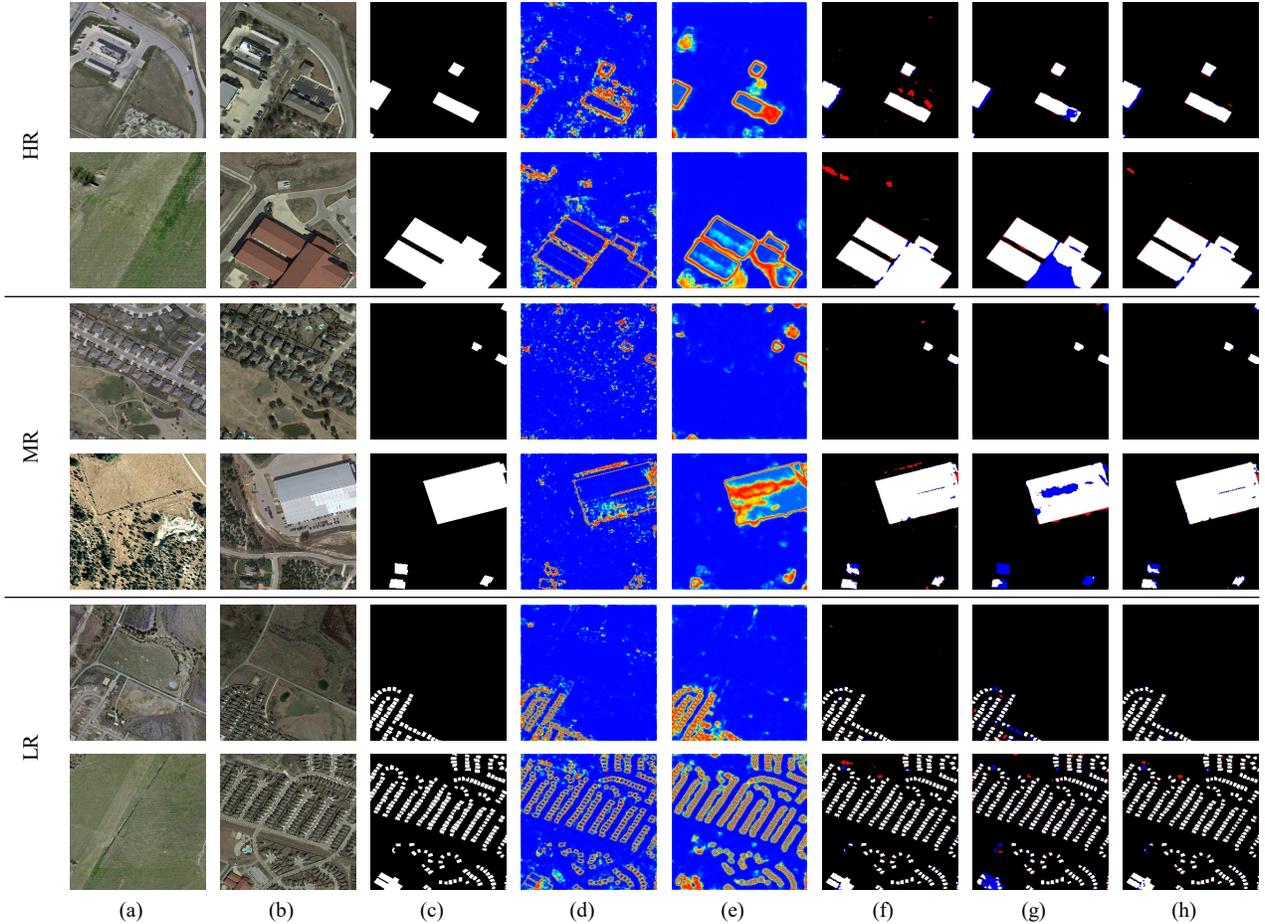

Fig. 10. Modules' output of WRICNet on HR, MR, and LR data set. (a) Image $T_1$. (b) Image $T_2$. (c) Ground Truth. (d) Heat map of weighted output WRI module. (e) Heat map of weighted output of WRC module. (f) Output of WRI module. (g) Output of WRC module. (h) Output of WRICNet.

## E. Ablation Study

To verify the effectiveness of our training strategy, and network architecture, we compare our WRICNet with its ablated versions:

(1) proposed: The proposed(WRICNet) in this article.
(2) w/o multi-channel: Changing the network input structure of the proposed from Concatenation to Siamese+Concatenation.
(3) w/o weighted class: In the training process, the appropriate weights are no longer assigned to the change, and non-change samples to solve the problem of unbalanced sample distribution in the data set. That is, the weight of the change, and non-change samples in the the training process is set to 1:1.
(4) w/o weighted scale block: Removing the weighted scale block of the WRI module, and WRC module.
(5) w/o inception module v2: Removing the improvement of the Inception module in this article.
(6) w/o rich-scale block: Replacing the Rich-scale block with a single Conv block-3.
(7) w/o rich-scale block v2: Removing the improvement of the Rich-scale block in this article.

In the ablation study, the parameters of the model, the average values of the LOI, and GI on the HR, MR, and LR data set are illustrated in Table V.

TABLE V
ABLATIONS STUDY OF PROPOSED ON HR, MR, AND LR DATA SET

| Method | parameters | Local Optimal Index | | | | | | | | Global Index | | | |
|---|---|---|---|---|---|---|---|---|---|---|---|---|---|
| | | 5% | | | | 10% | | | | | | | |
| | | MA | FA | F1 | IoU | MA | FA | F1 | IoU | MA | FA | F1 | IoU |
| proposed | 2,902,357 | 6.0 | 5.9 | **94.1** | **88.8** | 7.3 | **6.1** | **93.3** | **87.5** | 18.3 | **10.3** | **85.4** | **74.7** |
| w/o multi-channel | 2,896,932 | 6.1 | 6.6 | 93.6 | 88.0 | 7.4 | 6.9 | 92.9 | 86.7 | **17.4** | 12.0 | 85.2 | 74.2 |
| w/o weighted class | 2,902,357 | 6.5 | 6.2 | 93.7 | 88.1 | 8.1 | 6.7 | 92.6 | 86.2 | 19.6 | 11.9 | 84.0 | 72.4 |
| w/o weighted scale block | 2,591,077 | 6.4 | **5.8** | 93.7 | 88.5 | 7.3 | 6.5 | 92.9 | 87.1 | 18.3 | 11.1 | 84.9 | 73.9 |
| w/o inception module v2 | 3,833,813 | 5.9 | 6.8 | 93.6 | 88.0 | 7.2 | 7.3 | 92.7 | 86.5 | 19.4 | 12.3 | 83.8 | 72.4 |
| w/o rich-scale block | 3,706,491 | 7.7 | 7.9 | 92.3 | 85.7 | 8.8 | 8.5 | 91.3 | 84.2 | 20.6 | 13.4 | 82.8 | 70.8 |
| w/o rich-scale block v2 | 2,202,553 | **5.5** | 6.6 | 93.9 | 88.6 | **6.6** | 7.0 | 93.2 | 87.3 | 18.3 | 11.7 | 84.5 | 73.5 |

Through careful observation, we can find that some of the following rules:
(1) The proposed has achieved the best F1, and IoU in the top 5%, 10% LOI, and GI, respectively, 94.1%, 88.8%, and 93.3%, 87.5%, and 85.4%, 74.7%.
(2) Compared to w/o multi-channel, the proposed has slightly more parameters, whereas F1 has increased by 0.5%, 0.4%, and 0.2% in the top 5%, 10% LOI, and GI respectively, and IoU increased by 0.8%, 0.8%, and 0.5% in the top 5%, 10% LOI, and GI respectively. The above demonstrates that the Siamese structure can slightly reduce the model parameters, but in this article it fails to improve the CD effect.
(3) Compared to w/o weighted class, F1 of the proposed has increased by 0.4%, 0.7%, and 1.4% in the top 5%, 10% LOI, and GI respectively, and IoU increased by 0.7%, 1.3%, and 2.3% in the top 5%, 10% LOI, and GI respectively. It is proved that the training strategy of assigning appropriate weights to the change samples, and the non-change samples in the training process to achieve sample balance is effective.
(4) Compared to w/o weighted scale block, F1 of the proposed has increased by 0.4%, 0.4%, and 0.5% in the top 5%, 10% LOI, and GI respectively, and IoU increased by 0.3%, 0.4%, and 0.8% in the top 5%, 10% LOI, and GI respectively. It is proved that the Weighted scale block, which assigns appropriate weights to features of different scales, in this article, is effective.
(5) Compared to w/o inception module v2, F1 of the proposed has increased by 0.5%, 0.6%, and 1.6% in the top 5%, 10% LOI, and GI respectively, and IoU increased by 0.8%, 1.0%, and 2.3% in the top 5%, 10% LOI, and GI respectively. The proposed parameters are 32% less than the parameters of w/o inception module v2, and since reducing the number of parameters of the model speeds up the convergence of the model, thus, the CD effect is improved. All in all, it is effective for the improvement of the Inception module.
(6) Compared to w/o rich-scale block, F1 of the proposed has increased by 1.8%, 2.0%, and 2.6% in the top 5%, 10% LOI, and GI respectively, and IoU increased by 3.1%, 3.3%, and 3.9% in the top 5%, 10% LOI, and GI respectively. The proposed parameters are 27% less than the parameters of w/o rich-scale block. The above demonstrates that Rich-scale block can reduce model parameters, and improve the CD effect by extracting multi-scale features.
(7) Compared to w/o rich-scale block v2, F1 of the proposed has increased by 0.2%, 0.1%, and 0.9% in the top 5%, 10% LOI, and GI respectively, and IoU increased by 0.2%, 0.2%, and 1.2% in the top 5%, 10% LOI, and GI respectively. It proves that the improvement of the Rich-scale block in this article is effective. Although the model parameter amount is increased by 21%, the effect of CD is improved.

IV. CONCLUSION

With the purpose of making the model achieve great effect in multi-resolution RS CD data sets, we designs the Weighted scale block, which assigns appropriate weights to features of different scales, and strengthens the ability to express the edges of changing areas. WRI module is used to obtain SMF, which consists of Weighted scale block, Rich-scale block, and Inception module. WRC module is used to obtain DMF, which consists of Weighted scale block, Rich-scale block, and U-net. Then, fusing the extracted SMF, and DMF via Concatenate, and a Metric module is used to obtain the CD output. The performance experiment on the multi -resolution data set with comparative methods proves that the proposed(WRICNet) can complement SMF, and DMF well, reduce the MA in the change area, and FA outside the change area, which makes the edge of the change area more accurate. Therefore, CD outputs of proposed in HR , MR, and LR data set are closest to Ground Truth. The LOI ,GI, and model parameters of comparative methods, and the proposed proves the proposed can achieve better change detection effect on the basis of fewer model parameters. The heat map of weighted output of WRI module, WRC module, and output of WRI module, WRC module, WRICNet on HR, MR, and LR data set prove that WRI module, and WRC module can obtain SMF, and DMF respectively, Weighted scale block can make a great expression of the edge of the changing area, and WRICNet can complement SMF, and DMF well. At the end, the ablation study illustrates that the training strategy, the proposed Weighted scale block, and the improvements of the Rich-scale block, and Inception module are effective.
ACKNOWLEDGMENTACKNOWLEDGMENT

The authors would like to appreciate the authors of opening data sets for their contributions.

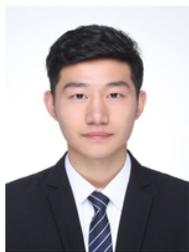
**Yu Jiang** was born in Jiangxi, China in 1996. He received the B.E. degree from the School of Computer Science, Minnan Normal University, Fujian, China. He is currently pursuing the M.E. degree the School of Computer Information Engineering, Jiangxi Normal University, Nanchang, China.

His research interests include deep learning and remote sensing image change detection.

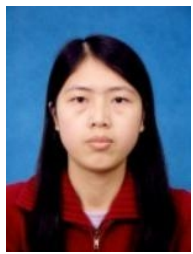
**Lei Hu** received her BS degree in computer software from Central China Normal University, China, in 2002, her MS degree in computer systems architecture from Huazhong University of Science and Technology, China, in 2005, and her PhD in computer application from Beihang University, China, in 2013. She is an associate professor at the School of Computer Information Engineering, Jiangxi Normal University, China.

Her current research interests include remote sensing image processing, pattern recognition, and machine learning.

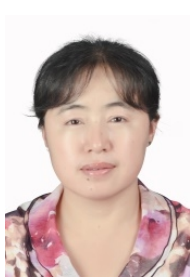
**Yongmei Zhang** received her BS degree from Peking University, China, in 1990 and her MSc and PhD degrees from North University of China, China, in 2002 and 2006, respectively. She did her postdoctoral research in Beihang University, China, in 2008. In 2012, she was a visiting scholar at Peking University. Currently, she works as a professor at the North China University of Technology, China. She is a member of CCF and ACM.

Her research interests include artificial intelligence, pattern recognition, and image processing.

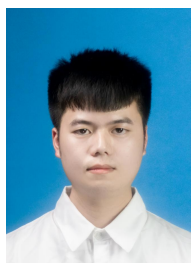
**Xin Yang** received the B.E. degree from Pingxiang University, Jiangxi, China, in 2019. He is currently pursuing the M.E. degree with Jiangxi Normal University, Nanchang, China.

His current research focuses on deep learning and remote sensing image change detection.